\documentclass[conference]{IEEEtran}
\IEEEoverridecommandlockouts
\usepackage{cite}
\usepackage{amsmath,amssymb,amsfonts}
\usepackage{algorithm}
\usepackage{graphicx}
\usepackage{textcomp}
\usepackage{xcolor}
\usepackage{url}

\usepackage{amsmath}
\DeclareMathOperator*{\argmax}{arg\,max}
\usepackage[ruled,vlined,algo2e]{algorithm2e}
\usepackage{algpseudocode}
\SetKwInput{KwIn}{Input}
\SetKwInput{KwOut}{Output}

\makeatletter
\newcommand{\rmnum}[1]{\romannumeral #1}
\newcommand{\Rmnum}[1]{\expandafter\@slowromancap\romannumeral #1@}
\makeatother

\def\BibTeX{{\rm B\kern-.05em{\sc i\kern-.025em b}\kern-.08em
    T\kern-.1667em\lower.7ex\hbox{E}\kern-.125emX}}
\begin{document}

\title{FreeGraftor: Training-Free Cross-Image Feature Grafting for Subject-Driven Text-to-Image Generation}

% \author{Anonymous ICME submission}
\author{
    \IEEEauthorblockN{Zebin Yao$^{1}$, Lei Ren$^{2}$, Huixing Jiang$^{2}$, Wei Chen$^{2}$, Xiaojie Wang$^{1}$, Ruifan Li$^{1}$, Fangxiang Feng$^{1}$}
    \IEEEauthorblockA{$^1$ College of Artificial Intelligence, Beijing University of Posts and Telecommunications, Beijing, China}
    \IEEEauthorblockA{$^2$ Li Auto Inc., Beijing, China}
    \IEEEauthorblockA{\{zebin.yao, xjwang, rfli, fxfeng\}@bupt.edu.cn, \{renlei3, jianghuixing, chenwei10\}@lixiang.com}
}

\maketitle

\begin{abstract}
Subject-driven image generation aims to synthesize novel scenes that faithfully preserve subject identity from reference images while adhering to textual guidance. However, existing methods struggle with a critical trade-off between fidelity and efficiency. Tuning-based approaches rely on time-consuming and resource-intensive subject-specific optimization, while zero-shot methods fail to maintain adequate subject consistency. In this work, we propose FreeGraftor, a training-free framework that addresses these limitations through cross-image feature grafting. Specifically, FreeGraftor leverages semantic matching and position-constrained attention fusion to transfer visual details from reference subjects to the generated images. Additionally, our framework introduces a novel noise initialization strategy to preserve the geometry priors of reference subjects, facilitating robust feature matching. Extensive qualitative and quantitative experiments demonstrate that our method enables precise subject identity transfer while maintaining text-aligned scene synthesis. Without requiring model fine-tuning or additional training, FreeGraftor significantly outperforms existing zero-shot and training-free approaches in both subject fidelity and text alignment. Furthermore, our framework can seamlessly extend to multi-subject generation, making it practical for real-world deployment. Our code is available at \url{https://github.com/Nihukat/FreeGraftor}.
\end{abstract}
\section{Introduction}
With the rapid advancement of diffusion models\cite{rombach2022high,podell2023sdxl,esser2024scaling,flux2024}, text-to-image technology has evolved significantly, enabling photorealistic and diverse imagery generation from textual descriptions. Beyond generic synthesis, it has spawned personalized content creation tasks, among which subject-driven text-to-image generation is critical. This task aims to synthesize novel scenes with specific subjects via reference images and text prompts, requiring simultaneous high-fidelity subject preservation and precise text alignment.

Existing subject-driven methods face an inherent trade-off between fidelity and efficiency. These methods can be roughly divided into two categories based on whether test-time fine-tuning is required. One is \textbf{tuning-based approaches} (e.g., DreamBooth\cite{ruiz2023dreambooth}), which adapt models through subject-specific fine-tuning on reference images. They achieve strong subject consistency but have notable drawbacks, including high computational cost, time consumption, catastrophic forgetting, and poor multi-subject handling. The other is \textbf{zero-shot methods} (e.g., IP-Adapter\cite{ye2023ip}), which skip fine-tuning. Instead, they use auxiliary modules to extract and inject subject features from references. They offer high efficiency but fail to retain fine details, resulting in unsatisfactory alignment.

\begin{figure}[t!]
\centering
\includegraphics[width=0.8\linewidth]{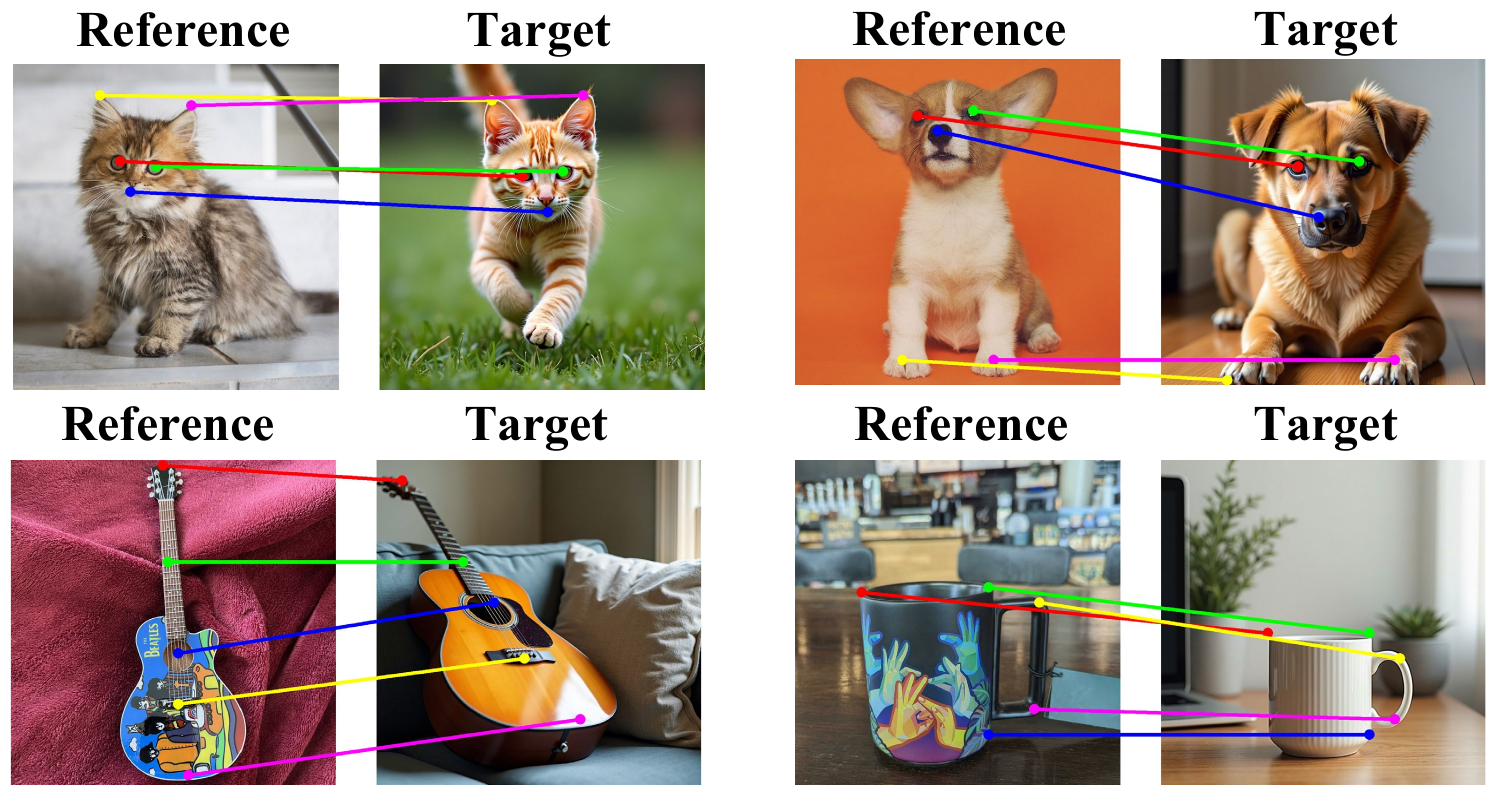}
\caption{Cross-image semantic correspondence in FLUX.1. Semantically corresponding pixels between reference and generated images show high similarity in the model's feature space.}
\label{fig:semantic_correspondence}
\end{figure}

We find the key to resolving this dilemma lies in the powerful visual representation of pretrained text-to-image models. Specifically, robust cross-image semantic correspondences exist in modern MM-DiT-based models (e.g., FLUX.1\cite{flux2024}). As shown in Fig. \ref{fig:semantic_correspondence}, we compute token-level cosine similarities between the reference and target images in FLUX.1 during diffusion, linking the most similar pixel pairs. This visualization confirms that semantically corresponding regions have high feature similarity in FLUX.1, providing a reliable foundation for training-free subject feature transfer.

Leveraging this, we propose FreeGraftor, a training-free framework the resolves the fidelity-efficiency dilemma via cross-image feature grafting. It extracts reference features via rectified flow inversion and injects them dynamically during generation. Specifically, we design a novel \textbf{cross-image feature grafting} mechanism transferring reference details to semantically matched regions, integrating semantic matching (for feature patch alignment) and position-constrained attention fusion (for reliable transfer). Additionally, we propose a \textbf{structure-consistent initialization} strategy which involves constructing a collage containing the subject and inverting it as generation initialization noise to ensure structural preservation. FreeGraftor achieves pixel-level detail retention with fast inference and low overhead, requiring no extra training.

Extensive experiments demonstrate that our method outperforms existing approaches in both subject consistency and text alignment while maintaining competitive efficiency. Notably, it scales well to multi-subject scenarios, preserving individual identities and integrating subjects harmoniously into coherent scenes without layout conflicts or attribute confusion, expanding practical applications of subject-driven generation.

Our contributions are summarized as follows:  

\begin{itemize}
    \item We introduce FreeGraftor, a plug-and-play framework for subject-driven image generation that achieves pixel-level detail preservation and flexible text-guided control without fine-tuning or training.  
    \item We develop a cross-image feature grafting technique that transfers visual characteristics of reference subjects to semantically corresponding regions in generated images via semantic matching and position-constrained attention fusion.  
    \item We propose a novel noise initialization strategy that preserves reference geometry through automated collage construction and rectified flow inversion, seamlessly extending our method to multi-subject scenarios.  
\end{itemize}

\section{RELATED WORK}

\subsection{Text-to-Image Diffusion Models}
Text-to-image diffusion models\cite{rombach2022high,podell2023sdxl,esser2024scaling,flux2024} have achieved remarkable success in generating high-quality and diverse images. Early popular open-source models, such as Stable Diffusion\cite{rombach2022high} and SDXL\cite{podell2023sdxl}, are based on the U-Net architecture and utilize cross-attention between image and text to incorporate textual conditions. In contrast, Diffusion Transformer (DiT)\cite{peebles2023scalable} replaces the U-Net backbone with a Transformer, demonstrating superior scalability and performance. Recent Multimodal-Diffusion Transformers (MM-DiTs), including Stable Diffusion 3\cite{esser2024scaling} and FLUX.1\cite{flux2024}, follow this design and project text and image tokens into a unified space to compute joint attention, thereby better integrating complex relationships between text and images. 

\subsection{Subject-Driven Generation}
Subject-driven generation aims to synthesize novel scenes containing user-specified subjects. Mainstream approaches fall into two categories: tuning-based\cite{ruiz2023dreambooth, kumari2023multi, gu2023mix} and zero-shot\cite{li2023blip, ye2023ip, wu2025less, mou2025dreamo}. \textbf{Tuning-based methods}(e.g., Dreambooth\cite{ruiz2023dreambooth}) require per-subject fine-tuning to capture reference subjects’ visual details. This pipeline is resource-heavy and slow, risks overfitting and catastrophic forgetting, and performs poorly in multi-subject composition. \textbf{Zero-shot methods} introduce additional learnable modules to process extra image inputs, enabling personalization at inference time. Early methods (e.g., IP-Adapter\cite{ye2023ip}) rely on ViT-based external encoders to compress images into embeddings, achieving only semantic-level alignment and suffering from insufficient visual consistency, particularly for novel concepts. Recent works (e.g., UNO\cite{wu2025less}) leverage the VAEs within pretrained text-to-image models to encode reference images for better subject consistency, yet they still struggle to retain fine-grained visual details (e.g., text and patterns).

Beyond these categories, several \textbf{training-free methods}\cite{ding2024freecustom, shin2024large, kang2025flux} harness pre-trained models' in-context learning ability for plug-and-play subject-driven generation. FreeCustom\cite{ding2024freecustom} concatenates reference images and noise in latent space via self-attention weighted masking to focus on reference subjects. DiptychPrompting\cite{shin2024large} extends to MM-DiT models via ControlNet\cite{zhang2023adding}, reframing the task as diptych image completion with tailored prompting. LatentUnfold\cite{kang2025flux} dispenses with ControlNet and replaces diptych with grid for better consistency. These methods avoid heavy training but are limited by base models' in-context learning capability, failing to fully reconstruct reference details via self-attention alone. Our method identifies cross-image semantic correspondences, binds matched patches, and reinforces reference-generated contextual connections to ensure full component consistency.

% \subsection{Feature Injection Techniques}
% Feature injection methods transfer visual characteristics from reference images to generated content. Cross-image attention \cite{alaluf2024cross} and feature matching \cite{mou2025dreamomatcher} have been used to align reference and generated features, but these approaches often require additional training or struggle with semantic consistency.

% In contrast to existing work, FreeGraftor achieves pixel-level detail preservation through semantic matching and position-constrained attention fusion without requiring any training or fine-tuning, resolving the critical trade-off between fidelity and efficiency in subject-driven generation.
\section{METHOD}
\subsection{Overview}
\begin{figure}[t!]
\centering
\includegraphics[width=0.95\linewidth]{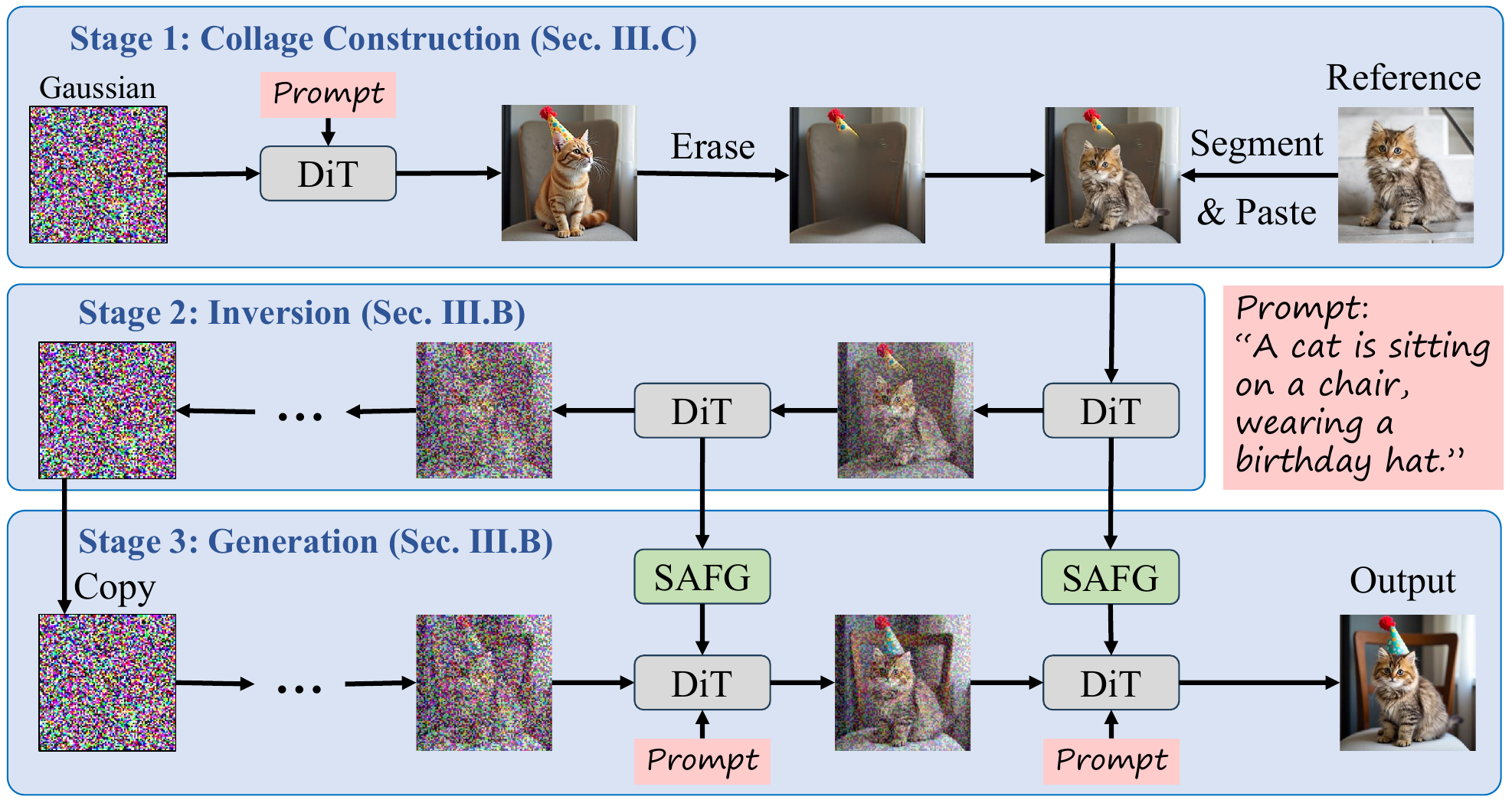}
\caption{Overview of FreeGraftor. The framework consists of three stages: (1) Collage construction using reference image and text prompt, (2) Inversion to obtain initialization noise, and (3) Generation with feature grafting.}
\label{fig:framework}
\end{figure}

Our framework comprises three stages: collage construction, inversion, and generation (Fig. \ref{fig:framework}). First, we construct a collage using the given reference image and text prompt. Next, we invert the collage to obtain initialization noise for the generation phase while extracting reference features. Finally, these features are injected into the generation process to synthesize the final image that contains the reference subject within the text-described scene.

\subsection{Semantic-Aware Feature Grafting}
Most subject-driven generation methods rely on learnable parameters to inject reference features into the generation process. Tuning-based approaches embed subject-specific information into pre-trained models, while zero-shot methods employ additional modules for feature extraction and injection—both incurring significant computational overhead.

\begin{figure}[t!]
\centering
\includegraphics[width=0.9\linewidth]{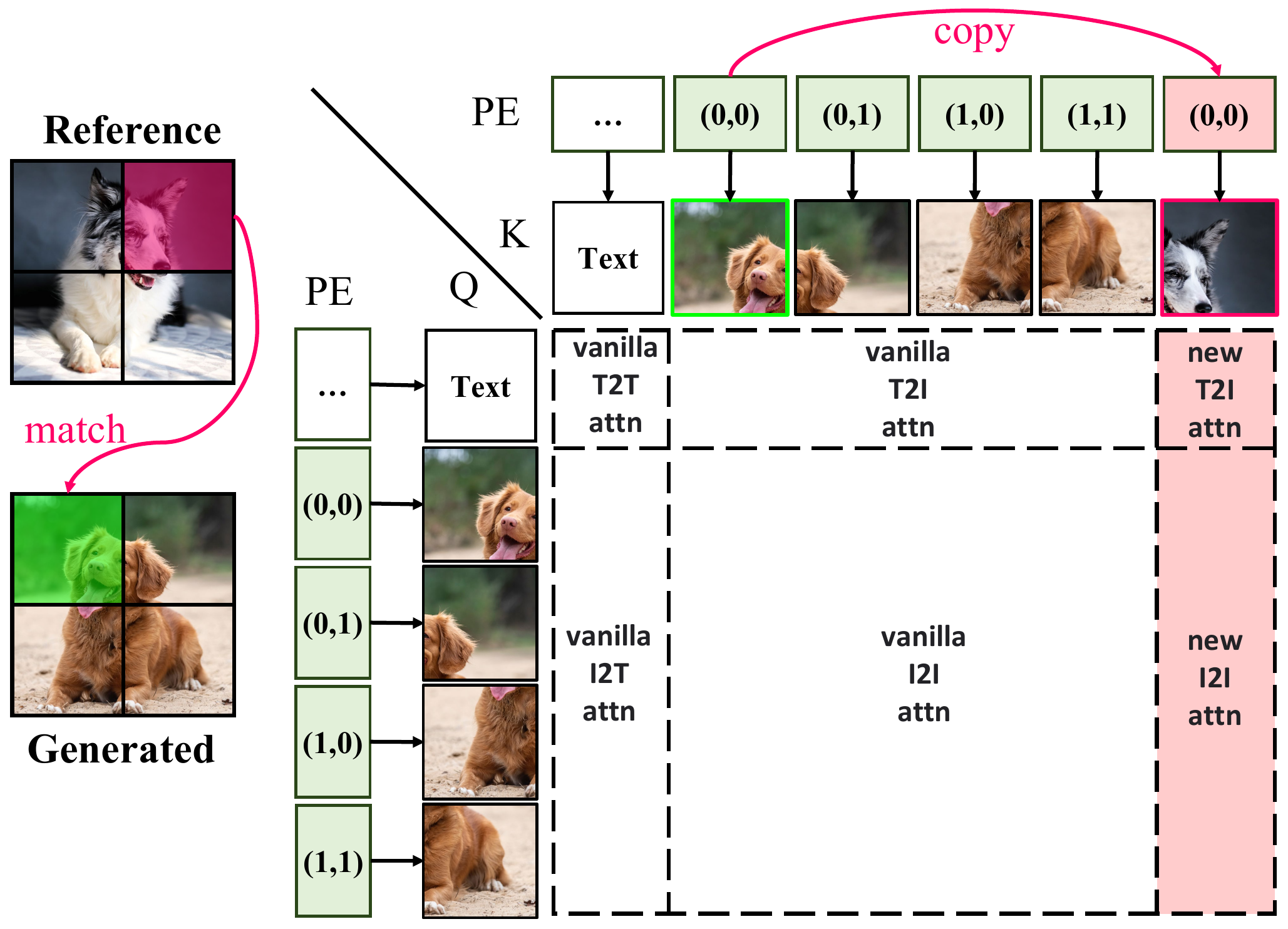}
\caption{Semantic-Aware Feature Grafting (SAFG) Module. The module establishes semantic correspondence and fuses features through position-constrained attention.}
\label{fig:safg_module}
\vspace{-0.2\baselineskip}
\end{figure}

Inspired by appearance transfer techniques\cite{tang2023emergent,go2024eye}, we leverage the inherent capabilities of text-to-image diffusion models to transfer visual characteristics of reference subjects to corresponding regions in generated images without incurring additional training costs. To achieve this, we propose the Semantic-Aware Feature Grafting (SAFG) module, which first establishes semantic correspondences between reference and generated patches via semantic matching, then fuses features through position-constrained attention fusion (Fig. \ref{fig:safg_module}).

\subsubsection{Semantic Matching}

Early appearance transfer methods directly concatenate keys and values of reference and generated images to compute cross-image attention. However, high attention scores do not necessarily indicate semantic correspondences across images, and erroneous matches may lead to structural distortions and visual artifacts. DIFT\cite{tang2023emergent} revealed that U-Net-based diffusion models contain rich semantic information in their feature space and established cross-image correspondences via cosine similarity-based semantic matching. We extend this principle to MM-DiT-based models.  

Given a reference image $I^{\text{ref}}$ and a predefined binary mask $M^{\text{ref, pre}}$ indicating the reference subject, we extract attention features using rectified flow inversion and match reference $H^{\text{ref}}$ and generated $H^{\text{gen}}$ features at the same timestep and attention layer. For each reference patch $i$ within $M^{\text{ref, pre}}$, we compute cosine similarity with all generated patches $j$:  

\begin{equation}
    s_{ij} = \frac{H_i^{\text{img,ref}} \cdot H_j^{\text{img,gen}}}{\|H_i^{\text{img,ref}}\|_2 \|H_j^{\text{img,gen}}\|_2}, \quad m(i) = \arg\max_j s_{ij}
\end{equation}

where $H_i^{\text{img,ref}}$ and $H_j^{\text{img,gen}}$ denote features of the $i$-th reference patch and $j$-th generated patch at the attention layer input. To ensure reliable matches, we apply two filtering strategies:  

\textbf{\rmnum{1}. Similarity Threshold Filtering.} For each reference patch $ i $, if its maximum cosine similarity $ s_{i, m(i)} $ falls below a predefined threshold $ \tau $, the match is discarded. This is represented by the binary mask: 

\begin{equation}
    M_i^{\text{ref, sim}} = \begin{cases}  
        1, & s_{i, m(i)} \geq \tau \\  
        0, & s_{i, m(i)} < \tau  
    \end{cases}
\end{equation}

\textbf{\rmnum{2}. Cycle Consistency Filtering.} Following DreamMatcher\cite{nam2024dreammatcher}, for each reference patch $i$ (satisfying $M_i^{\text{ref, pre}} = 1$), we find the most similar reference patch $k$ to its matched generated patch $m(i)$:  

\begin{equation}
    m^{-1}(m(i)) = \argmax_{\substack{k:  M_k^{\text{ref,pre}}=1}} 
    \frac{H_{m(i)}^{\text{img,gen}} \cdot H_k^{\text{img,ref}}}{\lVert H_{m(i)}^{\text{img,gen}} \rVert_2 \, \lVert H_k^{\text{img,ref}} \rVert_2}
\end{equation}

Let $\mathbf{p}_i^{\text{ref}}$ and $\mathbf{p}_{m^{-1}(m(i))}^{\text{ref}}$ denote the 2D coordinates of patch $i$ and its reverse-matched counterpart. Matches are rejected if their L2 distance exceeds $\delta$:  

\begin{equation}
    d_i = \|\mathbf{p}_i^{\text{ref}} - \mathbf{p}_{m^{-1}(m(i))}^{\text{ref}}\|_2, \quad  
    M_i^{\text{ref, consi}} = \begin{cases}  
        1, & d_i \leq \delta \\  
        0, & d_i > \delta  
    \end{cases}
\end{equation}

Combining these filters with the predefined mask yields the final reference image mask:

\begin{equation}
    M^{\text{ref}} = M^{\text{ref, pre}} \odot M^{\text{ref, sim}} \odot M^{\text{ref, consi}}
\end{equation}

\subsubsection{Position-Constrained Attention Fusion}

After obtaining the filtered reference mask $M^\text{ref}$, we transfer features from the retained reference patches. A naive approach is to directly replace the features of the corresponding patches. However, this may introduce artifacts along object boundaries or even cause image distortion (as shown in Fig. \ref{fig:ablation}). Unlike U-Net, which models positional relationships through convolutional operations, MM-DiT distinguishes patch proximity and orientation via position embeddings (PE). This enables us to align reference patches with their target positions in the generated image by editing position embeddings.  

Specifically, we concatenate both the keys and values of the reference patch to those of the generated patch, along with the position embeddings corresponding to the generated patch associated with the reference patch. This design enables the reference patch and its corresponding generated patch to share identical positional information. Let $ K^\text{txt} $ and $ V^\text{txt} $ denote the text embeddings' keys and values, respectively; $ K^\text{img, gen} $ and $ V^\text{img, gen} $ represent the generated image's keys and values; and $ K_{M}^\text{img, ref} $ and $ V_{M}^\text{img, ref} $ correspond to the keys and values of the reference patches satisfying $ M^\text{ref} $. We concatenate these as follows:  

\begin{equation}
\begin{aligned}
    K^\text{cat} = [K^\text{txt}; K^\text{img,gen}; K_M^\text{img,ref}], \\ 
    V^\text{cat} = [V^\text{txt}; V^\text{img,gen}; V_M^\text{img,ref}]
\end{aligned}
\end{equation}

For each retained reference patch $ i $, we extract the corresponding position  embedding $ PE_{m(i)}^\text{img, gen} $ based on the matching relationship $ m(i) $. The new position  embedding is then formed by concatenating the following embeddings:  

\begin{equation}
    PE^\text{cat} = [PE^\text{txt}; PE^\text{img,gen}; \{PE_{m(i)}^{\text{img,gen}} \mid M_i^\text{ref} = 1\}]
\end{equation}

The revised attention computation is formulated as:  

\begin{equation}
    \tilde{Q} = Q \circledast PE, \quad \tilde{K}^\text{cat} = K^\text{cat} \circledast PE^\text{cat}
\end{equation}

% \vspace{-1.2\baselineskip}

\begin{equation}
    A^+ = \text{softmax}\left(\frac{\tilde{Q} (\tilde{K}^\text{cat})^\top}{\sqrt{d}}\right), \quad H_{\text{out}}^+ = A^+ V^\text{cat}
\end{equation}

Here, $\circledast$ denotes the RoPE operation, $A^+$ represents the updated attention scores, and $H_{\text{out}}^+$ denotes the output of the attention layer. By integrating semantic matching and the sharing of position  embeddings, our SAFG module effectively binds features between matching patches in the reference and generated images.  

\subsection{Structure-Consistent Initialization}

Attention-based feature migration replicates the appearance of reference subjects (e.g., color, texture) but may not preserve structural details (e.g., shape, body proportions). Moreover, Gaussian noise initialization introduces significant randomness, leading to discrepancies in shape and size between generated subjects and those in the reference image, potentially causing semantic matching failures. To maintain structural integrity and enhance the robustness of semantic matching, we propose a structure-consistent initialization strategy. This involves creating a collage containing the reference subject and then employing an inversion technique to derive the initial noise required for the generation phase.  

As illustrated in Fig. \ref{fig:framework}, given a text prompt (e.g., "a cat sitting on a chair wearing a birthday hat"), we first use a base text-to-image model to generate a template image that faithfully represents the described scene. To preserve the structure of the reference subject, we replace the corresponding subject in the template image with the reference subject. Specifically, a grounding model localizes the target subject in the template, a segmentation model extracts its mask, and an inpainting model removes the subject from the template. The reference subject is then cropped (using the same grounding and segmentation models), resized, and pasted into the erased region to form a collage. This collage replaces the original reference image as the new reference. Finally, we invert this collage to obtain the initial noise for the generation phase and record the diffusion trajectory during inversion to facilitate feature grafting.

To prevent the generated subject from overfitting the pose of the reference subject, we introduce a dynamic feature dropout strategy. Since structural layouts are predominantly determined during early diffusion steps, we reduce the injection of reference features during the early steps while retaining them in the later steps to preserve visual details. At timestep $t$, for each reference patch, we apply dropout with probability $p = \omega t$, where $\omega \in [0, 1]$ controls dropout intensity. The dropout mask is represented by a Bernoulli distribution:  

\begin{equation}
    M^\text{drop}_{i,j} \sim \text{Bernoulli}(1 - \omega t) \quad \forall i, j
\end{equation}

The final reference mask is computed by applying dropout to the reference image mask obtained from semantic matching:  

\begin{equation}
    {M^{\text{ref}}}^{\prime} = M^\text{ref} \odot M^\text{drop}
\end{equation}

This strategy preserves the reference subject’s structure while aligning it with the text-driven layout and maintaining pose flexibility. For multi-subject generation, we iteratively replace subjects in the template and invert the multi-subject collage, avoiding simultaneous processing of multiple reference images and eliminating computational overhead.

\section{EXPERIMENTS}

\begin{figure*}[t!]
\centering
\includegraphics[width=0.95\linewidth]{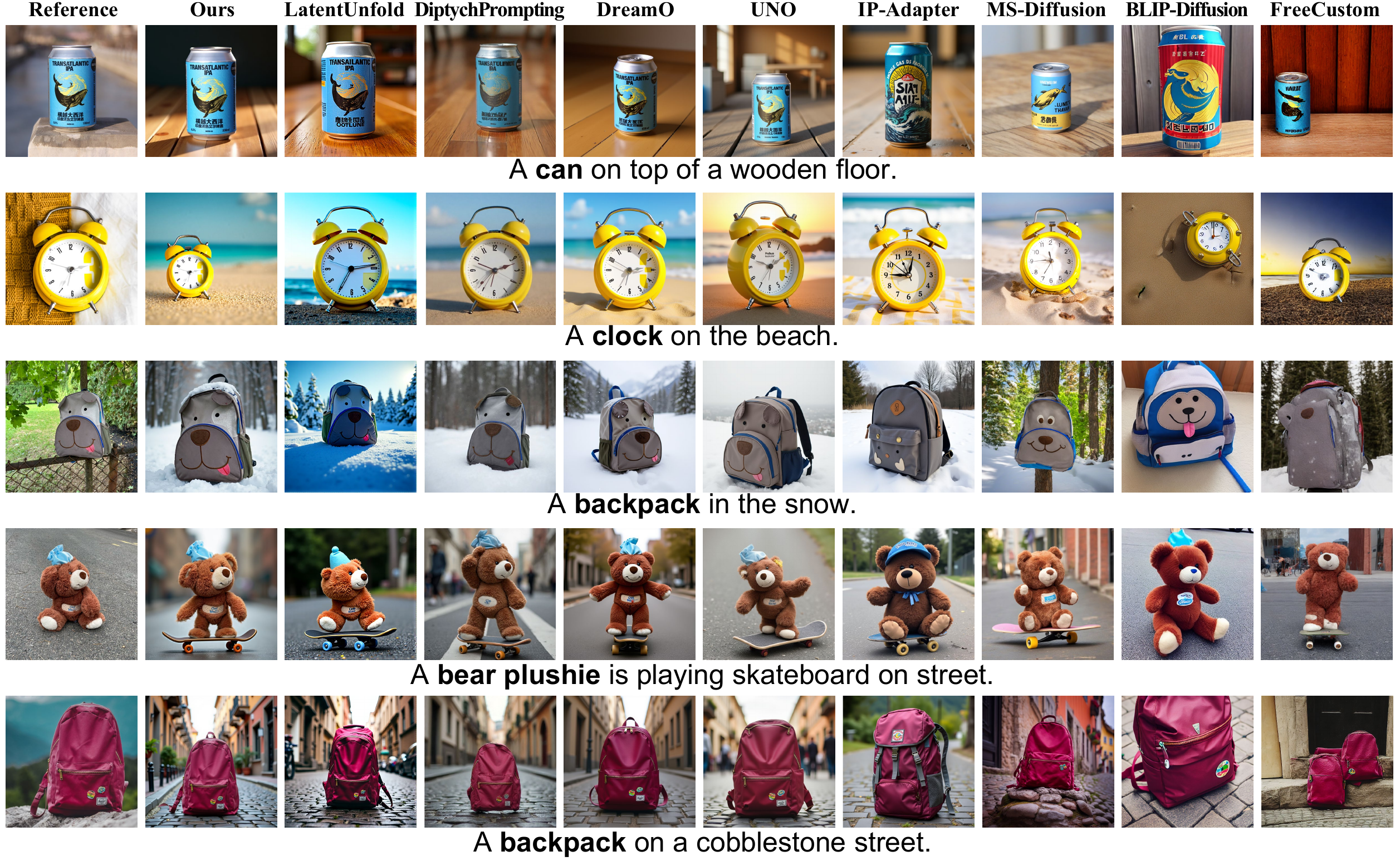}
\caption{Single-subject generation results. FreeGraftor preserves pixel-level details while enabling flexible text-guided control.}
\label{fig:qualitative_single}
\end{figure*}

\begin{figure}[t!]
\centering
\includegraphics[width=0.95\linewidth]{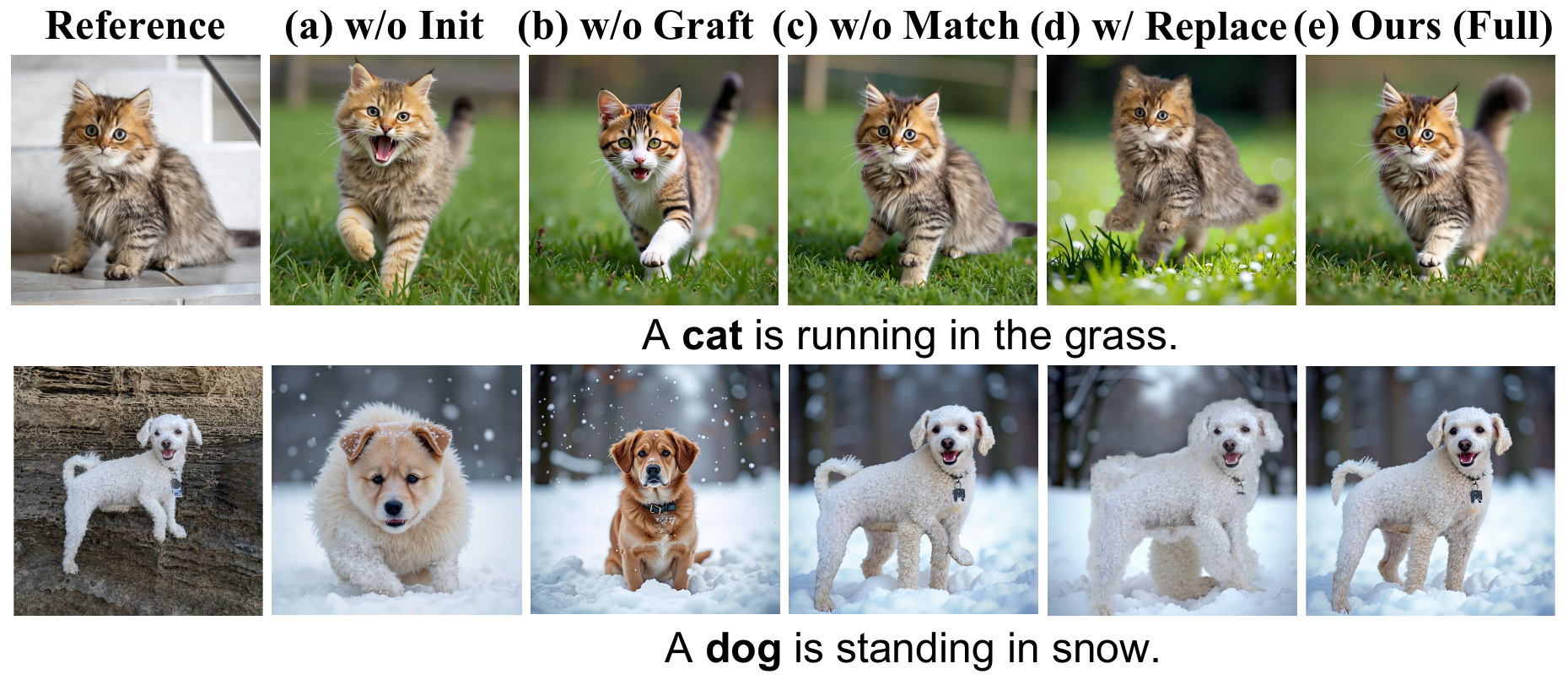}
\caption{Ablation study results. Each component of FreeGraftor is critical for achieving high-quality subject-driven generation.}
\label{fig:ablation}
% \vspace{-1.3\baselineskip}
\end{figure}

\subsection{Experimental Setup}

\subsubsection{Implementation Details}

Our method is implemented on FLUX.1-dev. For collage construction, we use Grounding DINO\cite{liu2024grounding} to localize subjects, SAM\cite{kirillov2023segment} to segment subjects, and LaMa\cite{suvorov2022resolution} to erase the original subject in the template image. FireFlow\cite{deng2024fireflow} is employed for rectified flow inversion. Template generation, inversion, and final generation each take 25 diffusion steps. During semantic matching, we set the similarity threshold $\tau = 0.2$, cycle consistency threshold $\delta = 1.5$, and dropout intensity $\omega = 0.5$.  

\subsubsection{Baselines}

We compare our method with zero-shot methods (BLIP-Diffusion\cite{li2023blip}, IP-Adapter\cite{ye2023ip}, MS-Diffusion\cite{wang2024ms}, UNO\cite{wu2025less}, DreamO\cite{mou2025dreamo}) and training-free methods (FreeCustom\cite{ding2024freecustom}, DiptychPrompting\cite{shin2024large}, LatentUnfold\cite{kang2025flux}). Details of these methods are provided in Appendix B. 

\subsubsection{Datasets and Evaluation}

For qualitative evaluation, we use reference images from DreamBench\cite{ruiz2023dreambooth}, CustomConcept101\cite{kumari2023multi}, and Mix-of-Show\cite{gu2023mix} for subject-driven generation. For quantitative evaluation, we generate 30×25×4=3,000 images using 30 subjects and 25 prompt templates from DreamBench\cite{ruiz2023dreambooth}, with 4 random seeds per subject-prompt pair. We evaluate performance from two perspectives:  text alignment and image alignment. For text alignment, we report results using CLIP\cite{radford2021learning} and ImageReward\cite{xu2024imagereward} metrics. For image alignment, we use GroundedSAM\cite{ren2024grounded} to extract segments of reference and generated subjects, respectively, and compute similarity scores between these segments using CLIP\cite{radford2021learning} and DINOv2\cite{oquab2023dinov2}.

\begin{table}[t!]
\renewcommand{\arraystretch}{1.3}
\caption{Comparing to other subject-driven generation methods, our method achieves the best performance.}
\label{tab:quantitative}
\centering
\small
\setlength{\tabcolsep}{0.7mm}
\begin{tabular}{llcccccc}
\hline & & \multicolumn{2}{c}{ Image-alignment } && \multicolumn{2}{c}{ Text-alignment } \\
\cline { 3 - 4 } \cline { 6 - 7 }  Method & Base & CLIP-I$\uparrow$  & DINO$\uparrow$ && CLIP-T$\uparrow$ & IR$\uparrow$ \\
\hline 
FreeCustom & SD1.5 & 0.8308 & 0.6107 && 0.3246 & 0.6223  \\
BLIP-Diffusion & SD1.5 & 0.8621 & 0.7364 && 0.2949 & 0.2360  \\
MS-Diffusion & SDXL & 0.9023 & 0.7977 && 0.3254 & 1.3405  \\
IP-Adapter & FLUX & 0.8920 & 0.7696 && 0.3048 & 0.7444  \\
UNO & FLUX & 0.9069 & 0.8237 && 0.3151 & 1.2788  \\
DreamO & FLUX & 0.9159 & 0.8400 && 0.3217 & 1.6234  \\
DiptychPrompting & FLUX & 0.8924 & 0.7971 && 0.3291 & 1.5728 \\
LatentUnfold & FLUX & 0.8741 & 0.7755 && 0.3170 & 1.3929 \\
\hline
Ours & FLUX & $\mathbf{0.9527}$ & $\mathbf{0.9042}$ && $\mathbf{0.3308}$ & $\mathbf{1.6481}$   \\
\hline
\end{tabular}
\vspace{-0.5\baselineskip}
\end{table}

\begin{table}[t!]
\renewcommand{\arraystretch}{1.3}
\caption{Comparing to ablation variants, our full method achieves the best performance.}
\label{tab:ablation}
\centering
\small
\setlength{\tabcolsep}{1.5mm}
\begin{tabular}{lcclccl}
\hline & \multicolumn{2}{c}{ Image-alignment } && \multicolumn{2}{c}{ Text-alignment } \\
\cline { 2 - 3 } \cline { 5 - 6 }  Variant & CLIP-I$\uparrow$  & DINO$\uparrow$ && CLIP-T$\uparrow$ & ImageReward$\uparrow$  \\
\hline w/o Init & 0.8529 & 0.7138 && 0.3236 & 1.6320  \\
w/o Graft & 0.8463 & 0.6572 && 0.3306 & 1.6434  \\
w/o Match & 0.9504 & 0.8974 && 0.3232 & 1.5258  \\
w/ replace & 0.8476 & 0.7010 && 0.3174 & 1.4551  \\
\hline Ours & $\mathbf{0.9527}$ & $\mathbf{0.9042}$ && $\mathbf{0.3308}$ & $\mathbf{1.6481}$   \\
\hline
\end{tabular}
\end{table}

\subsection{Quantitative Results}
As shown in Table \ref{tab:quantitative}, our method achieves the highest alignment across all metrics, demonstrating superior subject fidelity and textual adherence, validating its effectiveness for subject-driven generation.

\subsection{Qualitative Results}
Fig. \ref{fig:qualitative_single} presents the results for single-subject generation using FreeGraftor and other methods. FreeCustom, BLIP-Diffusion and IP-Adapter fail to retain fine visual details of the reference subject effectively. While other baselines partially recover reference characteristics, they struggle with fine-grained fidelity. FreeGraftor achieves pixel-level detail preservation, such as logos on the can, numbers on the clock, and text on the teddy bear. For non-rigid subjects (e.g., the teddy bear), FreeGraftor generates natural pose variations aligned with text guidance, demonstrating the flexibility and robustness of our method.

\subsection{Ablation Study}
We compare several ablation variants in Figure \ref{fig:ablation}. (a) When structure-consistent initialization is removed, the generated images inherit the reference subject’s appearance (e.g., colors, patterns) but fail to maintain structural consistency (e.g., distorted body proportions in dogs). (b) Without SAFG, the generated subjects lack identifiable correspondence with the reference subjects. (c) If semantic matching is omitted and reference keys and values are directly appended during attention computation, the generated subjects exhibit rigid poses and artifacts. (d) Replacing position-constrained attention fusion with direct feature replacement introduces incoherent edges and distortions. Quantitative results in Table \ref{tab:ablation} confirm that both structure-consistent initialization and SAFG significantly improve image alignment, as they are critical for preserving subject identity. Skipping semantic matching or positional fusion degrades text alignment due to reduced image quality.  

\section{Conclusion}

We present FreeGraftor, a novel training-free framework for subject-driven generation that bridges the critical gap between subject fidelity and computational efficiency. By leveraging SAFG, FreeGraftor seamlessly injects the appearance of reference subjects into generated scenes. Additionally, our Structure-Consistent Initialization strategy preserves reference geometry and enhances the robustness of semantic matching. Extensive experiments demonstrate that FreeGraftor achieves pixel-level detail preservation and flexible text-guided control while requiring no training or test-time optimization. Our framework can be seamlessly extended to multi-subject generation, faithfully retaining visual details of all reference subjects while incurring no additional computational overhead.

\section*{Acknowledgment}
The work is supported by Key R\&D and Achievement Transformation Program of Inner Mongolia Autonomous Region (No. 2026SYFHH0570), National Key R\&D Program of China (No.2020YFF0305302), and the Science and Technology Project of State Grid Corporation of China (No. 5700-202058480A-0-0-00).

\bibliographystyle{IEEEbib}
\bibliography{main}

@inproceedings{rombach2022high,
  title={High-resolution image synthesis with latent diffusion models},
  author={Rombach, Robin and Blattmann, Andreas and Lorenz, Dominik and Esser, Patrick and Ommer, Bj{\"o}rn},
  booktitle={Proceedings of the IEEE/CVF conference on computer vision and pattern recognition},
  pages={10684--10695},
  year={2022}
}

@article{podell2023sdxl,
  title={Sdxl: Improving latent diffusion models for high-resolution image synthesis},
  author={Podell, Dustin and English, Zion and Lacey, Kyle and Blattmann, Andreas and Dockhorn, Tim and M{\"u}ller, Jonas and Penna, Joe and Rombach, Robin},
  journal={arXiv preprint arXiv:2307.01952},
  year={2023}
}

@inproceedings{peebles2023scalable,
  title={Scalable diffusion models with transformers},
  author={Peebles, William and Xie, Saining},
  booktitle={Proceedings of the IEEE/CVF international conference on computer vision},
  pages={4195--4205},
  year={2023}
}

@inproceedings{esser2024scaling,
  title={Scaling rectified flow transformers for high-resolution image synthesis},
  author={Esser, Patrick and Kulal, Sumith and Blattmann, Andreas and Entezari, Rahim and M{\"u}ller, Jonas and Saini, Harry and Levi, Yam and Lorenz, Dominik and Sauer, Axel and Boesel, Frederic and others},
  booktitle={Forty-first international conference on machine learning},
  year={2024}
}

@misc{flux2024,
    author={Black Forest Labs},
    title={FLUX},
    year={2024},
    howpublished={\url{https://github.com/black-forest-labs/flux}},
}

@inproceedings{ruiz2023dreambooth,
  title={Dreambooth: Fine tuning text-to-image diffusion models for subject-driven generation},
  author={Ruiz, Nataniel and Li, Yuanzhen and Jampani, Varun and Pritch, Yael and Rubinstein, Michael and Aberman, Kfir},
  booktitle={Proceedings of the IEEE/CVF conference on computer vision and pattern recognition},
  pages={22500--22510},
  year={2023}
}

@inproceedings{kumari2023multi,
  title={Multi-concept customization of text-to-image diffusion},
  author={Kumari, Nupur and Zhang, Bingliang and Zhang, Richard and Shechtman, Eli and Zhu, Jun-Yan},
  booktitle={Proceedings of the IEEE/CVF conference on computer vision and pattern recognition},
  pages={1931--1941},
  year={2023}
}

@article{gu2023mix,
  title={Mix-of-show: Decentralized low-rank adaptation for multi-concept customization of diffusion models},
  author={Gu, Yuchao and Wang, Xintao and Wu, Jay Zhangjie and Shi, Yujun and Chen, Yunpeng and Fan, Zihan and Xiao, Wuyou and Zhao, Rui and Chang, Shuning and Wu, Weijia and others},
  journal={Advances in Neural Information Processing Systems},
  volume={36},
  pages={15890--15902},
  year={2023}
}

@article{li2023blip,
  title={Blip-diffusion: Pre-trained subject representation for controllable text-to-image generation and editing},
  author={Li, Dongxu and Li, Junnan and Hoi, Steven},
  journal={Advances in Neural Information Processing Systems},
  volume={36},
  pages={30146--30166},
  year={2023}
}

@article{ye2023ip,
  title={Ip-adapter: Text compatible image prompt adapter for text-to-image diffusion models},
  author={Ye, Hu and Zhang, Jun and Liu, Sibo and Han, Xiao and Yang, Wei},
  journal={arXiv preprint arXiv:2308.06721},
  year={2023}
}

@article{wang2024ms,
  title={Ms-diffusion: Multi-subject zero-shot image personalization with layout guidance},
  author={Wang, Xierui and Fu, Siming and Huang, Qihan and He, Wanggui and Jiang, Hao},
  journal={arXiv preprint arXiv:2406.07209},
  year={2024}
}

@article{wu2025less,
  title={Less-to-more generalization: Unlocking more controllability by in-context generation},
  author={Wu, Shaojin and Huang, Mengqi and Wu, Wenxu and Cheng, Yufeng and Ding, Fei and He, Qian},
  journal={arXiv preprint arXiv:2504.02160},
  year={2025}
}

@article{mou2025dreamo,
  title={Dreamo: A unified framework for image customization},
  author={Mou, Chong and Wu, Yanze and Wu, Wenxu and Guo, Zinan and Zhang, Pengze and Cheng, Yufeng and Luo, Yiming and Ding, Fei and Zhang, Shiwen and Li, Xinghui and others},
  journal={arXiv preprint arXiv:2504.16915},
  year={2025}
}

@inproceedings{ding2024freecustom,
  title={Freecustom: Tuning-free customized image generation for multi-concept composition},
  author={Ding, Ganggui and Zhao, Canyu and Wang, Wen and Yang, Zhen and Liu, Zide and Chen, Hao and Shen, Chunhua},
  booktitle={Proceedings of the IEEE/CVF Conference on Computer Vision and Pattern Recognition},
  pages={9089--9098},
  year={2024}
}

@article{shin2024large,
  title={Large-Scale Text-to-Image Model with Inpainting is a Zero-Shot Subject-Driven Image Generator},
  author={Shin, Chaehun and Choi, Jooyoung and Kim, Heeseung and Yoon, Sungroh},
  journal={arXiv preprint arXiv:2411.15466},
  year={2024}
}

@article{kang2025flux,
  title={Flux Already Knows--Activating Subject-Driven Image Generation without Training},
  author={Kang, Hao and Fotiadis, Stathi and Jiang, Liming and Yan, Qing and Jia, Yumin and Liu, Zichuan and Chong, Min Jin and Lu, Xin},
  journal={arXiv preprint arXiv:2504.11478},
  year={2025}
}

@inproceedings{zhang2023adding,
  title={Adding conditional control to text-to-image diffusion models},
  author={Zhang, Lvmin and Rao, Anyi and Agrawala, Maneesh},
  booktitle={Proceedings of the IEEE/CVF international conference on computer vision},
  pages={3836--3847},
  year={2023}
}

@article{tang2023emergent,
  title={Emergent correspondence from image diffusion},
  author={Tang, Luming and Jia, Menglin and Wang, Qianqian and Phoo, Cheng Perng and Hariharan, Bharath},
  journal={Advances in Neural Information Processing Systems},
  volume={36},
  pages={1363--1389},
  year={2023}
}

@article{go2024eye,
  title={Eye-for-an-eye: Appearance transfer with semantic correspondence in diffusion models},
  author={Go, Sooyeon and Choi, Kyungmook and Shin, Minjung and Uh, Youngjung},
  journal={arXiv preprint arXiv:2406.07008},
  year={2024}
}

@inproceedings{nam2024dreammatcher,
  title={Dreammatcher: appearance matching self-attention for semantically-consistent text-to-image personalization},
  author={Nam, Jisu and Kim, Heesu and Lee, DongJae and Jin, Siyoon and Kim, Seungryong and Chang, Seunggyu},
  booktitle={Proceedings of the IEEE/CVF Conference on Computer Vision and Pattern Recognition},
  pages={8100--8110},
  year={2024}
}

@article{deng2024fireflow,
  title={FireFlow: Fast Inversion of Rectified Flow for Image Semantic Editing},
  author={Deng, Yingying and He, Xiangyu and Mei, Changwang and Wang, Peisong and Tang, Fan},
  journal={arXiv preprint arXiv:2412.07517},
  year={2024}
}

@inproceedings{liu2024grounding,
  title={Grounding dino: Marrying dino with grounded pre-training for open-set object detection},
  author={Liu, Shilong and Zeng, Zhaoyang and Ren, Tianhe and Li, Feng and Zhang, Hao and Yang, Jie and Jiang, Qing and Li, Chunyuan and Yang, Jianwei and Su, Hang and others},
  booktitle={European Conference on Computer Vision},
  pages={38--55},
  year={2024},
  organization={Springer}
}

@inproceedings{kirillov2023segment,
  title={Segment anything},
  author={Kirillov, Alexander and Mintun, Eric and Ravi, Nikhila and Mao, Hanzi and Rolland, Chloe and Gustafson, Laura and Xiao, Tete and Whitehead, Spencer and Berg, Alexander C and Lo, Wan-Yen and others},
  booktitle={Proceedings of the IEEE/CVF international conference on computer vision},
  pages={4015--4026},
  year={2023}
}

@article{ren2024grounded,
  title={Grounded sam: Assembling open-world models for diverse visual tasks},
  author={Ren, Tianhe and Liu, Shilong and Zeng, Ailing and Lin, Jing and Li, Kunchang and Cao, He and Chen, Jiayu and Huang, Xinyu and Chen, Yukang and Yan, Feng and others},
  journal={arXiv preprint arXiv:2401.14159},
  year={2024}
}

@inproceedings{suvorov2022resolution,
  title={Resolution-robust large mask inpainting with fourier convolutions},
  author={Suvorov, Roman and Logacheva, Elizaveta and Mashikhin, Anton and Remizova, Anastasia and Ashukha, Arsenii and Silvestrov, Aleksei and Kong, Naejin and Goka, Harshith and Park, Kiwoong and Lempitsky, Victor},
  booktitle={Proceedings of the IEEE/CVF winter conference on applications of computer vision},
  pages={2149--2159},
  year={2022}
}

@inproceedings{radford2021learning,
  title={Learning transferable visual models from natural language supervision},
  author={Radford, Alec and Kim, Jong Wook and Hallacy, Chris and Ramesh, Aditya and Goh, Gabriel and Agarwal, Sandhini and Sastry, Girish and Askell, Amanda and Mishkin, Pamela and Clark, Jack and others},
  booktitle={International conference on machine learning},
  pages={8748--8763},
  year={2021},
  organization={PMLR}
}

@article{xu2024imagereward,
  title={Imagereward: Learning and evaluating human preferences for text-to-image generation},
  author={Xu, Jiazheng and Liu, Xiao and Wu, Yuchen and Tong, Yuxuan and Li, Qinkai and Ding, Ming and Tang, Jie and Dong, Yuxiao},
  journal={Advances in Neural Information Processing Systems},
  volume={36},
  year={2024}
}

@article{oquab2023dinov2,
  title={Dinov2: Learning robust visual features without supervision},
  author={Oquab, Maxime and Darcet, Timoth{\'e}e and Moutakanni, Th{\'e}o and Vo, Huy and Szafraniec, Marc and Khalidov, Vasil and Fernandez, Pierre and Haziza, Daniel and Massa, Francisco and El-Nouby, Alaaeldin and others},
  journal={arXiv preprint arXiv:2304.07193},
  year={2023}
}

\clearpage
\appendix
\subsection{Pseudo Code}
\renewcommand{\thesection}{\Alph{section}.\arabic{section}}

\subsubsection{FreeGraftor Pipeline}
The complete FreeGraftor pipeline is outlined in Algorithm \ref{alg:pipeline_revised}, detailing the three main stages: collage construction, inversion, and generation.

\begin{algorithm}
\caption{FreeGraftor Pipeline}
\label{alg:pipeline_revised}
\SetKwProg{Fn}{Function}{}{}
\KwIn{Reference image $I^{\text{ref}}$, text prompt $P$}
\KwOut{Generated image $I^{\text{gen}}$}

\textbf{Stage 1: Collage Construction} \\
1. \textbf{Template Generation} \\
\quad $I^{\text{tmp}} \leftarrow \text{T2I}(P)$ \tcp*{Base text-to-image model} 

2. \textbf{Template Subject Processing} \\
\quad $B^{\text{tmp}} \leftarrow \text{GroundingDINO}(I^{\text{tmp}}, P)$ \tcp*{Bounding box} 
\quad $M^{\text{tmp}} \leftarrow \text{SAM}(I^{\text{tmp}}, B^{\text{tmp}})$ \tcp*{Segmentation mask} 
\quad $I^{\text{erase}} \leftarrow \text{LaMa}(I^{\text{tmp}}, M^{\text{tmp}})$ \tcp*{Inpainting} 

3. \textbf{Reference Subject Extraction} \\
\quad $B^{\text{ref}} \leftarrow \text{GroundingDINO}(I^{\text{ref}}, \text{"subject"})$ \\
\quad $M^{\text{ref}} \leftarrow \text{SAM}(I^{\text{ref}}, B^{\text{ref}})$ \\
\quad $I^{\text{crop}} \leftarrow \text{CropWithMask}(I^{\text{ref}}, M^{\text{ref}})$ \tcp*{Mask-based cropping} 

4. \textbf{Collage Assembly} \\
\quad $I^{\text{collage}} \leftarrow \text{ResizeAndPaste}(I^{\text{erase}}, I^{\text{crop}}, B^{\text{tmp}})$ \tcp*{Scale \& align} 

\textbf{Stage 2: Inversion} \\
5. $\epsilon^{\text{init}}, \mathcal{F}^{\text{ref}} \leftarrow \text{FireFlow}(I^{\text{collage}})$ \tcp*{Noise \& feature extraction} 

\textbf{Stage 3: Generation} \\
6. $I^{\text{gen}} \leftarrow \text{SAFG}(\epsilon^{\text{init}}, \mathcal{F}^{\text{ref}}, P)$ \tcp*{Feature grafting} 
\Return{$I^{\text{gen}}$}
\end{algorithm}

\subsubsection{Semantic-Aware Feature Grafting}
Algorithm \ref{alg:safg} details the Semantic-Aware Feature Grafting (SAFG) module, including semantic matching and position-constrained attention fusion.

\begin{algorithm}
\caption{Semantic-Aware Feature Grafting}
\label{alg:safg}
\KwIn{Reference features $\mathcal{F}^{\text{ref}}$, noise $\epsilon^{\text{init}}$, text prompt $P$}
\KwOut{Modified attention features $H_{\text{out}}^+$}

\textbf{Semantic Matching} \\
1. For each timestep $t$: \\
\quad \ForEach{reference patch $i \in M^{\text{ref, pre}}$}{
        \ForEach{generated patch $j$}{
            $s_{ij} \leftarrow \frac{H_i^{\text{img,ref}} \cdot H_j^{\text{img,gen}}}{\|H_i^{\text{img,ref}}\|_2\|H_j^{\text{img,gen}}\|_2}$ \\
            $m(i) \leftarrow \arg\max_j s_{ij}$
        }
    } 
2. Apply filtering: \\
\quad $M^{\text{ref}} \leftarrow \text{ThresholdFilter}(s_{ij}, \tau) \odot \text{CycleConsistency}(d_i, \delta)$ \\

\textbf{Position-Constrained Fusion} \\
3. Concatenate keys/values: \\
\quad $K^{\text{cat}} \leftarrow [K^{\text{txt}}; K^{\text{img,gen}}; K_M^{\text{img,ref}}]$ \\
\quad $V^{\text{cat}} \leftarrow [V^{\text{txt}}; V^{\text{img,gen}}; V_M^{\text{img,ref}}]$ \\
4. Bind positional  embeddings: \\
\quad $PE^{\text{cat}} \leftarrow \text{Concat}(PE^{\text{txt}}, PE^{\text{img,gen}}, \{PE_{m(i)}^{\text{img,gen}}\})$ \\
5. Compute revised attention: \\
\quad $\tilde{Q} \leftarrow PE \odot Q$ \\
\quad $\tilde{K}^{\text{cat}} \leftarrow PE^{\text{cat}} \odot K^{\text{cat}}$ \\
\quad $A^+ \leftarrow \text{Softmax}(\tilde{Q}(\tilde{K}^{\text{cat}})^\top/\sqrt{d})$ \\
\quad $H_{\text{out}}^+ \leftarrow A^+ V^{\text{cat}}$ \\
\Return{$H_{\text{out}}^+$}
\end{algorithm}
\subsection{Baseline Methods}
\renewcommand{\thesection}{\Alph{section}.\arabic{section}}

\subsubsection{Zero-Shot Methods}
Zero-shot methods do not require any subject-specific training or fine-tuning, making them efficient but often less effective at preserving subject details.
\begin{itemize}

\item\textbf{BLIP-Diffusion}\cite{li2023blip} trains a multimodal encoder to produce a unified text-aligned visual representation of the target subject. During training, it learns to leverage this subject representation so the diffusion model can generate new renditions of the subject from text prompts.

\item\textbf{IP-Adapter}\cite{ye2023ip} introduces a lightweight module that projects visual features from a CLIP image encoder directly into the cross-attention layers of a pre-trained text-to-image diffusion model. This enables the seamless integration of image prompts with textual conditions without the need for extra fine-tuning.

\item\textbf{MS-Diffusion}\cite{wang2024ms} injects reference images and explicit grounding tokens using a CLIP-based feature resampler to preserve fine details for each reference subject. Using the provided layout as guidance, it applies a multi-subject cross-attention mechanism so that each subject's conditioning affects a specific region of the image, which harmonizes the composition of all subjects while respecting the text prompt.

\item\textbf{UNO}\cite{wu2025less} first uses the inherent in-context generation ability of MM-DiTs to synthesize large-scale, high-consistency multi-subject paired data. Then it trains a "subject-to-image" model (conditioned on multiple images) with progressive cross-modal alignment and a universal rotary positional embedding. This pipeline yields a model that maintains high consistency while supporting control in both single- and multi-subject generation tasks.

\item\textbf{DreamO}\cite{mou2025dreamo} is a unified image customization framework built on a MM-DiT. DreamO is trained on a large mixed dataset covering diverse tasks (identity, subject, style, background, etc.), and uses techniques to integrate multiple conditions. It employs a feature-routing constraint that queries the relevant information from each reference, and a placeholder strategy that assigns each condition a dedicated position in the input. DreamO can perform many customization tasks with high quality and flexibly combine different types of control conditions in one model.

\end{itemize}

\subsubsection{Training-Free Methods}
Training-free methods avoid any form of training or fine-tuning, operating directly on pre-trained models.

\begin{itemize}

\item\textbf{FreeCustom}\cite{ding2024freecustom} provides a tuning-free approach for generating multi-concept images by leveraging a multi-reference self-attention mechanism along with a weighted mask strategy so that the generated image attends to each reference concept. With its weighted masking strategy to suppress extraneous details, this design efficiently preserves the semantic content of each input concept within context without further training.

\item\textbf{DiptychPrompting}\cite{shin2024large} reinterprets subject-driven generation as a text-conditioned inpainting task using a diptych layout, where the left panel holds a reference subject (with background removed) and the right panel is generated from a textual prompt. By enhancing the cross-attention between the two halves, DiptychPrompting precisely aligns the generated subject with the reference.

\item\textbf{LatentUnfold}\cite{kang2025flux} reframes subject generation as grid-based image completion: the reference images are simply tiled in a mosaic layout to "activate" the model's latent identity features. This simple "free lunch" approach is enhanced with a cascade attention design and a meta-prompting technique to boost fidelity.

\end{itemize}
\subsection{Additional Experiments}

\subsubsection{Hyperparameter Analysis}
We analyze the impact of key hyperparameters on performance, including consistency thresholds and dropout intensity.

\begin{figure}[!t]
\centering
\includegraphics[width=0.9\linewidth]{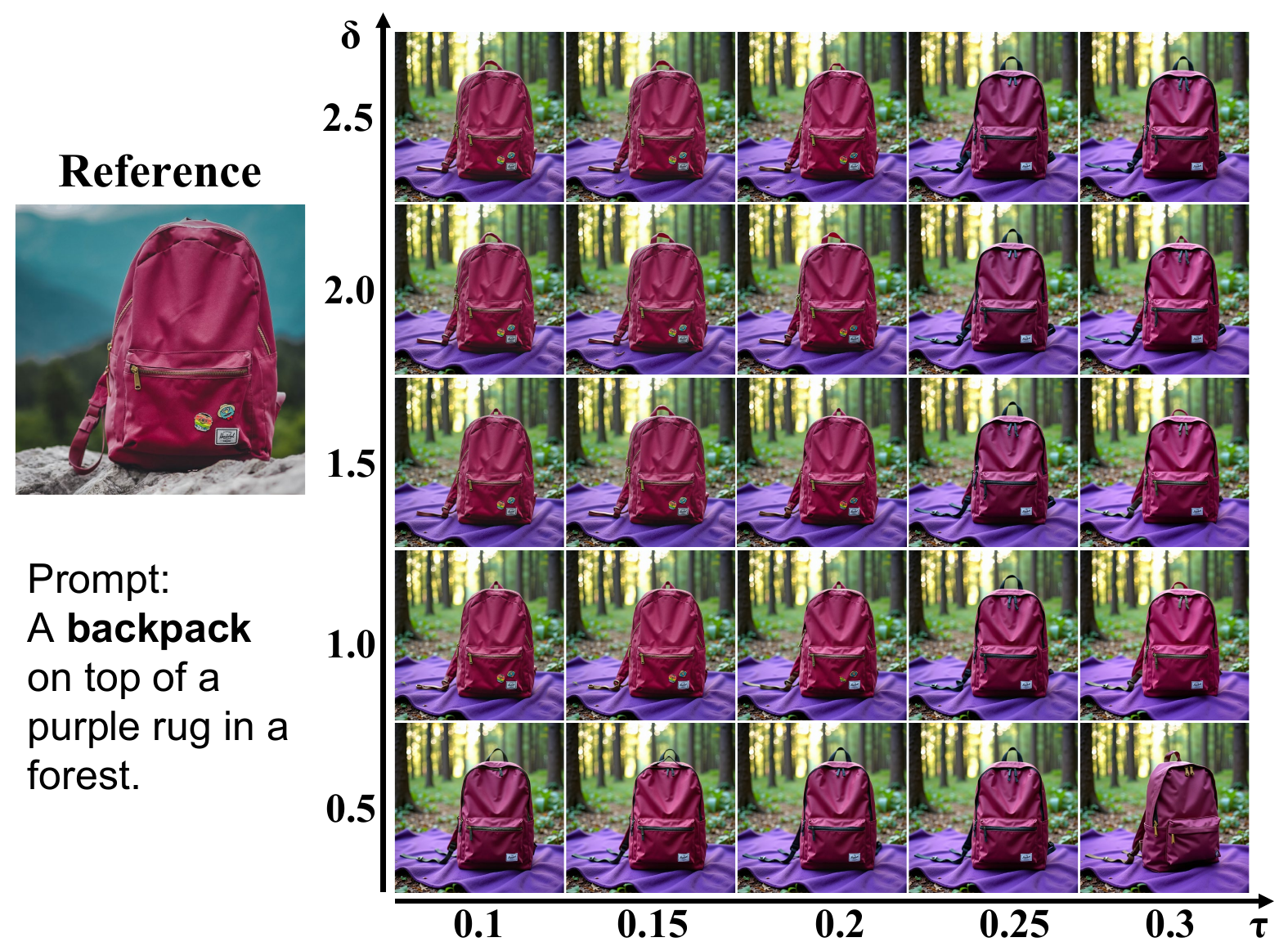}
\caption{Generation results under different similarity threshold ($\tau$) and cycle-consistency threshold ($\delta$) settings. Higher $\tau$ and lower $\delta$ lead to more stringent filtering.}
\label{fig:threshold_analysis}
\end{figure}

% \subsubsubsection{Similarity and Cycle Consistency Thresholds}
Fig. \ref{fig:threshold_analysis} shows generation results under different similarity threshold $\tau$ and cycle consistency threshold $\delta$ settings. Appropriate thresholds reduce mismatches (e.g., unnatural expansion of the backpack), while overly strict thresholds (higher $\tau$  or lower $\delta$) discard critical visual details (e.g., missing badges on the backpack).

\begin{figure}[!t]
\centering
\includegraphics[width=0.9\linewidth]{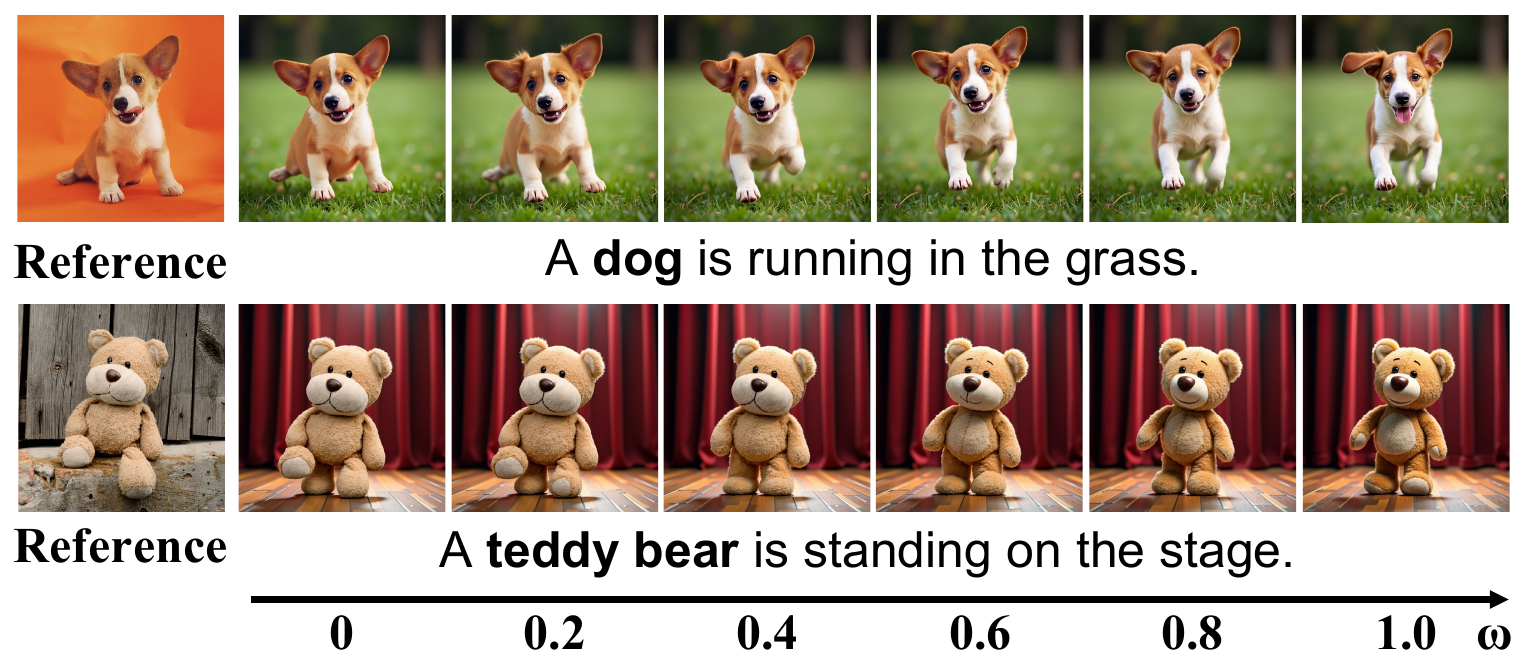}
\caption{Generation results under different dropout intensity ($\omega$) settings. Moderate dropout balances pose flexibility and detail preservation.}
\label{fig:dropout_analysis}
\end{figure}

% \subsubsubsection{Dropout Intensity}
Fig. \ref{fig:dropout_analysis} shows the effect of dropout intensity $\omega$. Moderate dropout ($\omega = 0.5$) balances pose flexibility and subject identity preservation. Excessive dropout ($\omega > 0.6$) degrades identity retention, while insufficient dropout restricts pose variation.

\subsubsection{Multi-Subject Generation Results}
\begin{figure*}[!t]
\centering
\includegraphics[width=0.95\linewidth]{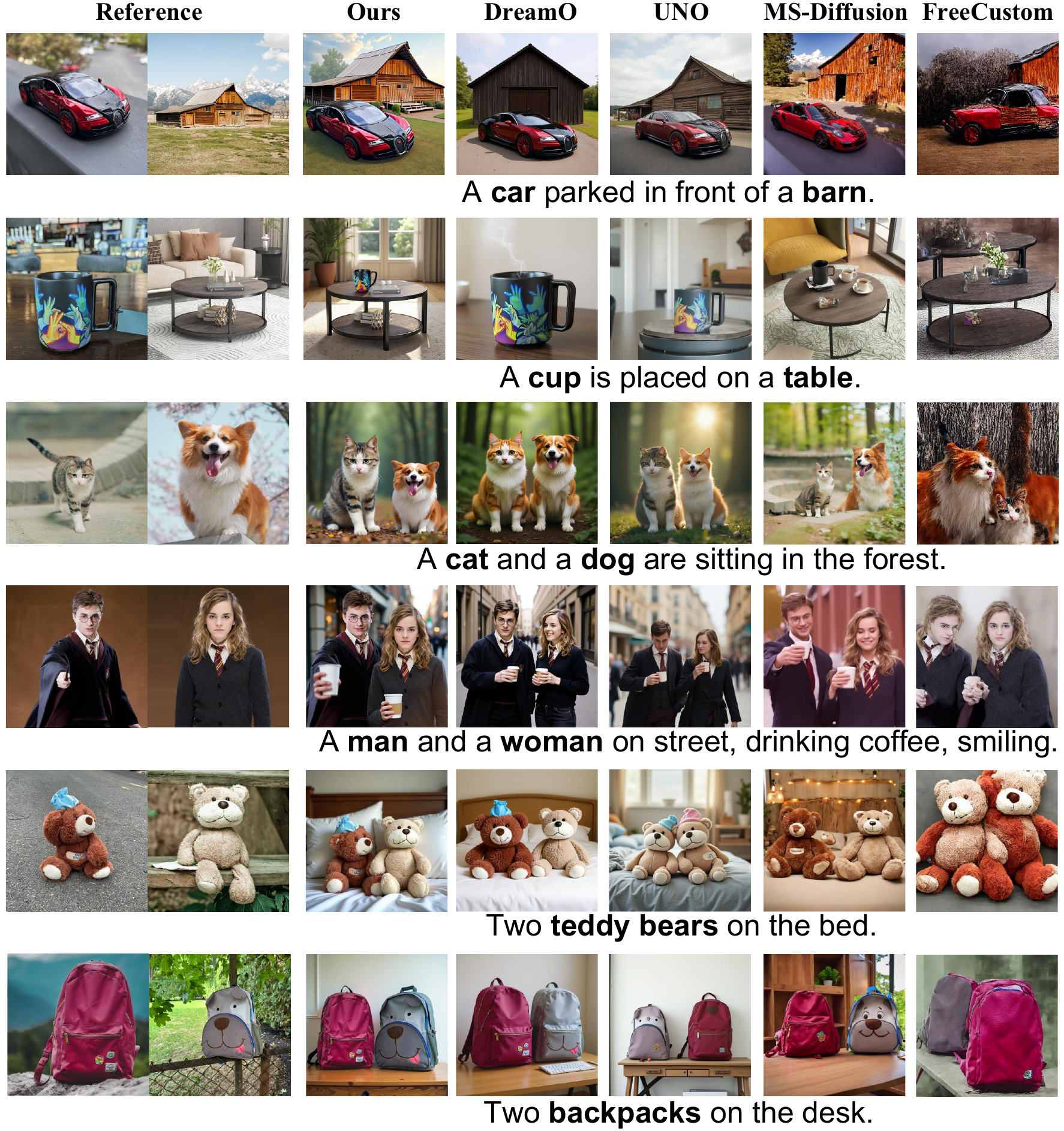}
\caption{Multi-subject generation results. FreeGraftor preserves all reference details without layout conflicts or attribute confusion.}
\label{fig:qualitative_multi}
\end{figure*}

FreeGraftor can seamlessly extend to multi-subject generation. As shown in Fig. \ref{fig:qualitative_multi}, FreeCustom and MS-Diffusion struggle to achieve sufficient subject consistency. While UNO and DreamO demonstrate higher fidelity, they suffer from subject omission (e.g., the barn and the table) and attribute confusion (e.g., confusion between the cat and the dog). FreeGraftor preserves visual details of all reference subjects while producing high-quality images well-aligned with the text, highlighting its effectiveness in multi-subject generation.  

\subsubsection{Efficiency Analysis}
Table \ref{tab:efficiency} compares the efficiency of FreeGraftor with other methods on an NVIDIA L20 GPU. All methods generate 512×512 resolution images under identical conditions.

\begin{table}[!t]
\renewcommand{\arraystretch}{1.3}
\small
\setlength{\tabcolsep}{0.7mm}
\centering
\caption{Efficiency Comparison of Subject-Driven Generation Methods}
\label{tab:efficiency}
\centering
\begin{tabular}{llrr}
% \toprule
\hline
Method & Base Model & Time (s) & Memory (MiB) \\
% \midrule
\hline
FreeCustom       & Stable Diffusion v1.5 & 14.93  & 12068.73 \\
BLIP-Diffusion   & Stable Diffusion v1.5 & 1.44   & 3641.84 \\
MS-Diffusion     & SDXL                  & 4.02   & 12268.41 \\
IP-Adapter       & FLUX.1-dev            & 7.45   & 38690.51 \\
UNO              & FLUX.1-dev            & 14.02  & 35248.43 \\
DreamO           & FLUX.1-dev            & 12.15  & 34431.61 \\
DiptychPrompting & FLUX.1-dev            & 43.27  & 44928.55 \\
LatentUnfold     & FLUX.1-dev            & 104.90 & 35402.22 \\
\hline
Ours             & FLUX.1-dev            & 40.93  & 28993.33 \\
% \bottomrule
\hline
\end{tabular}
\end{table}

As shown in Table \ref{tab:efficiency}, among methods built on the FLUX.1-dev base model, FreeGraftor significantly outperforms all other approaches in terms of memory usage, while also achieving shorter inference time compared to the other two training-free methods (DiptychPrompting and LatentUnfold). Compared to zero-shot methods, our approach is also competitive. While IP-Adapter, UNO, and DreamO have faster inference speeds than FreeGraftor, they consume more GPU memory. Due to their smaller parameter sizes, FreeCustom, BLIP Diffusion, and MS-Diffusion achieve high efficiency in both time and memory, but this comes at the cost of reduced subject consistency and lower visual quality. Notably, our FreeGraftor maintains an optimal balance between generation quality and computational efficiency, delivering robust performance without compromising subject faithfulness or visual details.

\subsubsection{Comparison with DreamBooth}
We compare FreeGraftor with DreamBooth, a widely-used tuning-based method. We implemented DreamBooth using LoRA (rank 16) on FLUX.1-Dev due to computational constraints.

\begin{figure*}[!t]
\centering
\includegraphics[width=0.95\linewidth]{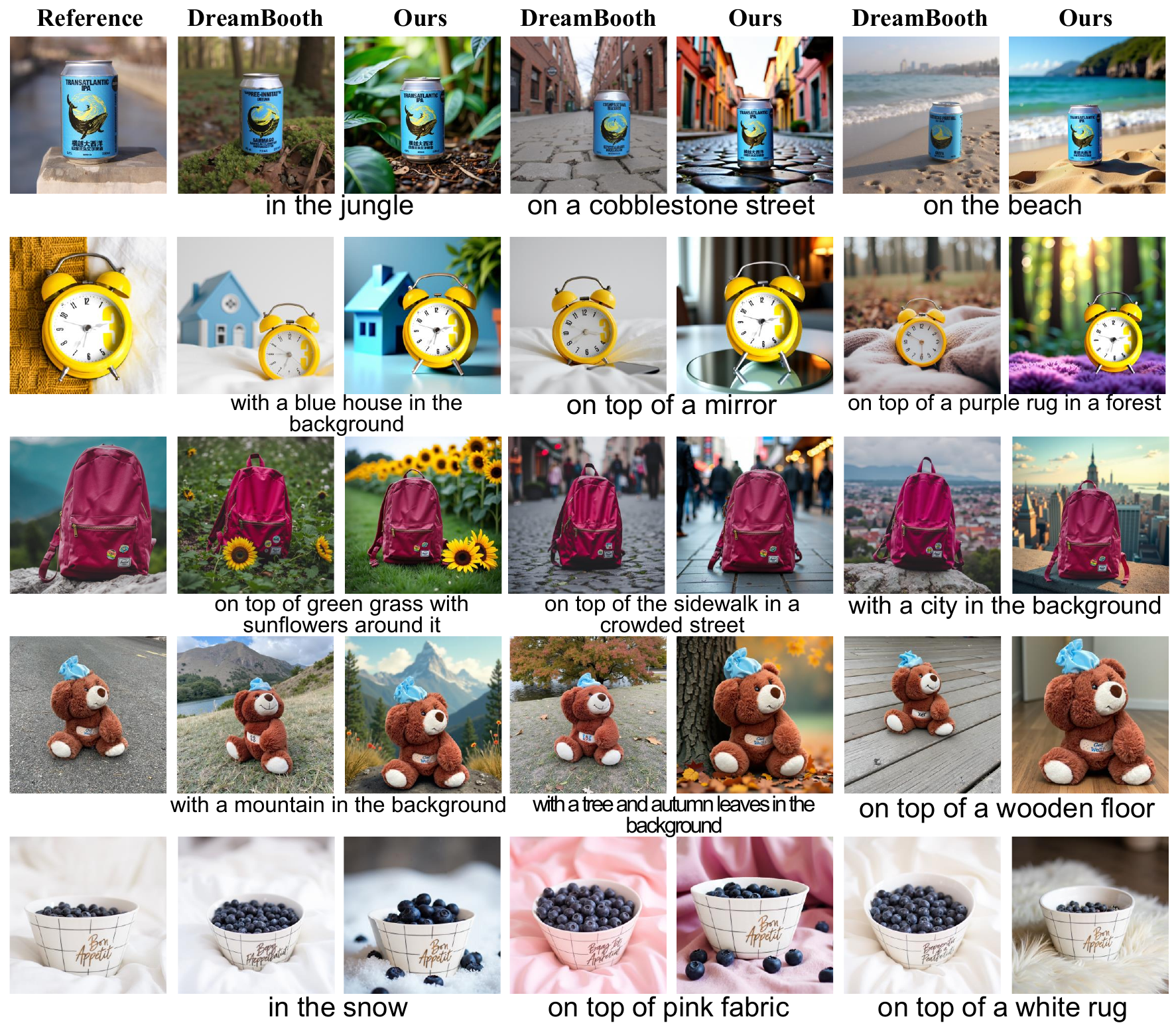}
\caption{Comparison between DreamBooth and our method on FLUX.1. Without computation-intensive fine-tuning, our method achieves superior detail preservation.}
\label{fig:dreambooth_comparison}
\end{figure*}

Fig. \ref{fig:dreambooth_comparison} shows qualitative results. Although fine-tuned for tens of minutes on each subject's reference images, DreamBooth struggles to fully reproduce visual details (e.g., text and patterns) and can overfit to background elements, compromising text alignment. In contrast, FreeGraftor achieves pixel-level detail preservation and faithful text alignment without resource-intensive fine-tuning.

% Table \ref{tab:dreambooth} shows quantitative results, demonstrating our significant performance advantages beyond just efficiency gains.

% \begin{table}[!t]
% \renewcommand{\arraystretch}{1.3}
% \caption{Quantitative Comparison with DreamBooth on FLUX.1-Dev}
% \label{tab:dreambooth}
% \centering
% \begin{tabular}{llcccc}
% \hline
% Method & Base Model & CLIP-I $\uparrow$ & DINO $\uparrow$ & CLIP-T $\uparrow$ & ImageReward $\uparrow$ \\
% \hline
% DreamBooth (LoRA) & FLUX.1-dev & 0.9412 & 0.8876 & 0.3267 & 1.6234 \\
% Ours & FLUX.1-dev & \textbf{0.9527} & \textbf{0.9042} & \textbf{0.3308} & \textbf{1.6481} \\
% \hline
% \end{tabular}
% \end{table}

\subsubsection{Comparison with DreamMatcher}
The most comparable approach to ours is likely DreamMatcher\cite{nam2024dreammatcher}, yet the two remain fundamentally distinct. Firstly, DreamMatcher is not a standalone subject-driven generation method but rather a plugin. It requires integration with tuning-based approaches (e.g., DreamBooth), which involve time-consuming and resource-intensive test-time optimization. In contrast, our method is a standalone subject-driven generation pipeline without any fine-tuning or training. Secondly, DreamMatcher adopts a direct attention feature replacement mechanism for identity injection, which tends to introduce artifacts along subject boundaries. In contrast, our approach leverages cross-image attention with shared spatial information, employing a more refined methodology that effectively mitigates image distortion.

\begin{figure*}[!t]
\centering
\includegraphics[width=0.95\linewidth]{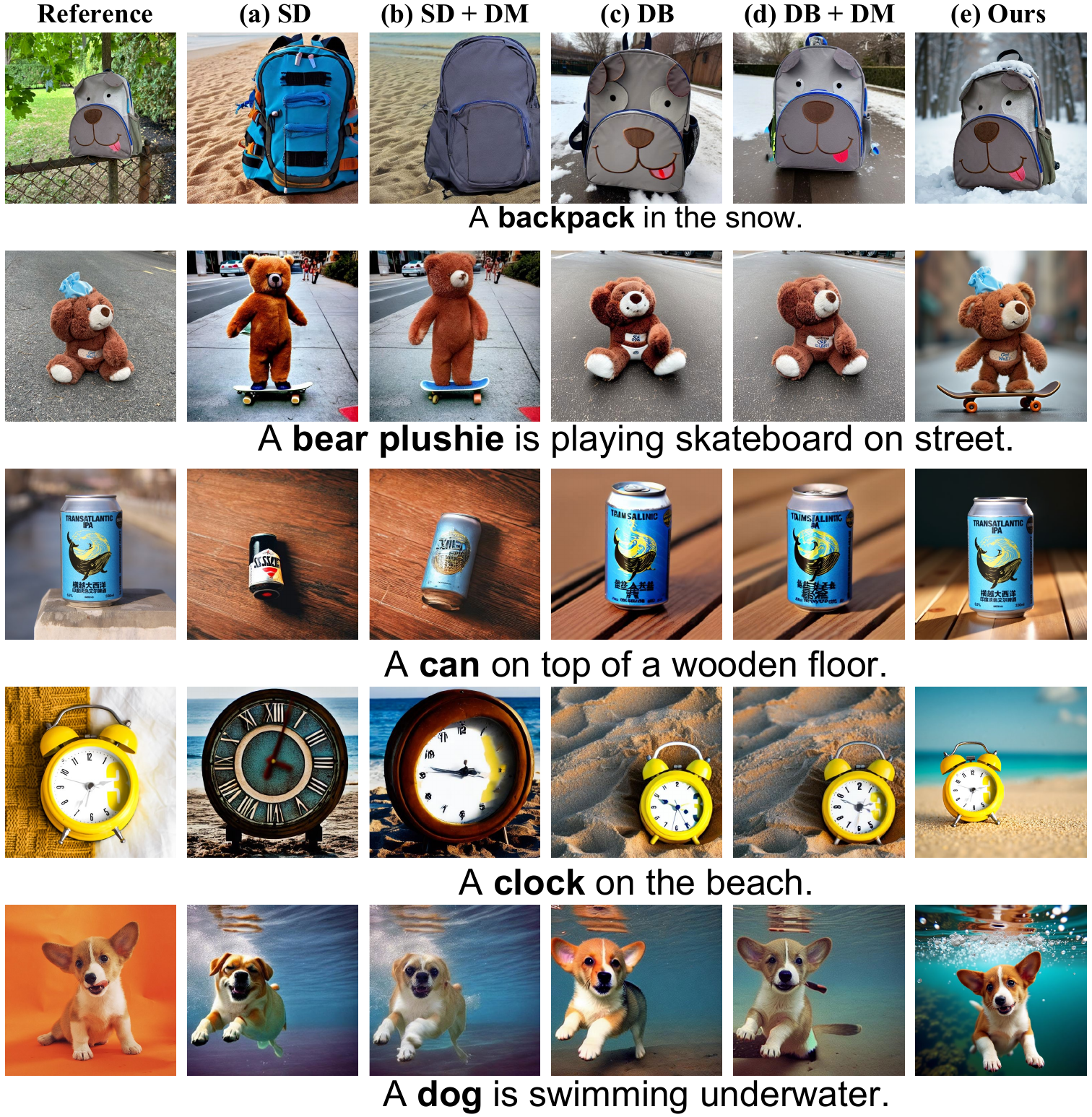}
\caption{Generation results of (a) Stable Diffusion (b) Stable Diffusion with DreamMatcher (c) DreamBooth on Stable Diffusion (d) DreamBooth on Stable Diffusion with DreamMatcher (e) Ours.}
\label{fig:dreammatcher_comparison}
\end{figure*}

Since DreamMatcher is tailored specifically for Stable Diffusion 1.5 and cannot be directly migrated to other network architectures, we implemented DreamMatcher on Stable Diffusion 1.5 following the official code. As show in Fig. \ref{fig:dreammatcher_comparison}, when employed in conjunction with DreamBooth, DreamMatcher effectively rectifies erroneous attribute binding(e.g., the yellow number "3" on the clock), thereby enhancing the fidelity of reconstructed visual details. However, when applied directly to Stable Diffusion, DreamMatcher is unable to fully reconstruct the subject from the reference image, owing to a lack of prior structural knowledge of the target subject. Additionally, some images generated by DreamMatcher exhibited noticeable distortions (e.g., green and blue artifacts around the backpack).

% Table \ref{tab:dreammatcher} confirms our superiority through both structure-consistent initialization and position-constrained attention fusion.

% \begin{table}[!t]
% \renewcommand{\arraystretch}{1.3}
% \caption{Quantitative Comparison with DreamMatcher}
% \label{tab:dreammatcher}
% \centering
% \begin{tabular}{llcccc}
% \hline
% Method & Base Model & CLIP-I $\uparrow$ & DINO $\uparrow$ & CLIP-T $\uparrow$ & ImageReward $\uparrow$ \\
% \hline
% SD + DreamMatcher & SD 1.5 & 0.7830 & 0.5336 & 0.3232 & 0.6387 \\
% DreamBooth + DreamMatcher & SD 1.5 & 0.8912 & 0.7918 & 0.3021 & 0.1767 \\
% Ours & FLUX.1-dev & \textbf{0.9527} & \textbf{0.9042} & \textbf{0.3308} & \textbf{1.6481} \\
% \hline
% \end{tabular}
% \end{table}

\subsubsection{Prompt Adaptability Results}
\begin{figure*}[!t]
\centering
\includegraphics[width=0.95\linewidth]{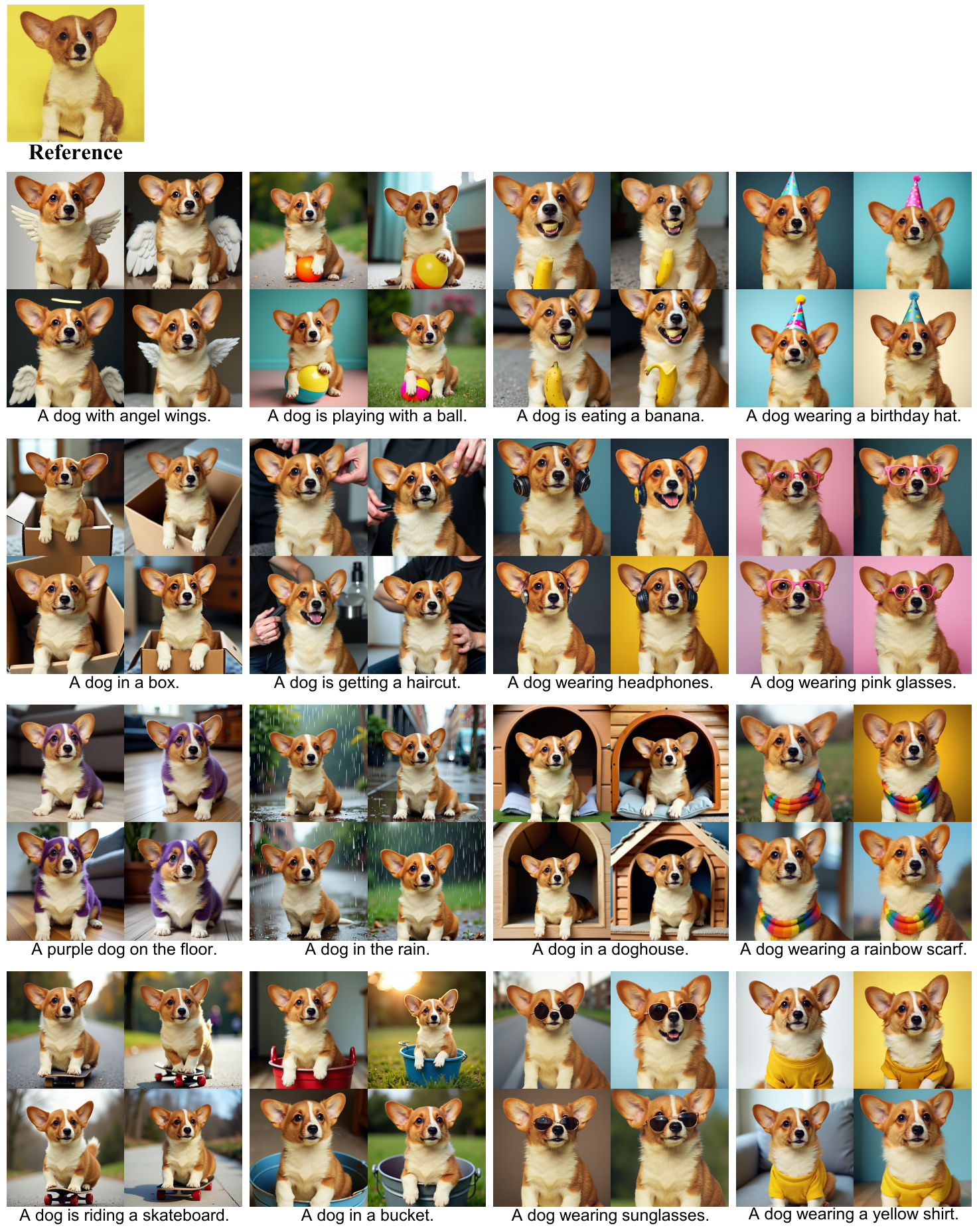}
\caption{Extended single-subject generation results with various prompts. FreeGraftor demonstrates robust prompt adaptability while preserving subject identity.}
\label{fig:prompt_adaptability}
\end{figure*}

Fig. \ref{fig:prompt_adaptability} demonstrates FreeGraftor's comprehensive prompt adaptability. Our method dynamically modulates subject attributes, poses, and scene configurations while maintaining pixel-level detail preservation, achieving both visual harmony and precise semantic alignment with input textual descriptions.

\subsection{Social Impact}

The proposed FreeGraftor framework has significant potential to democratize personalized content creation by enabling users without professional technical backgrounds to synthesize high-fidelity images of specific subjects without extensive training. This low-threshold advantage eliminates costly computing resource barriers that long excluded non-experts, empowering creators in digital art, advertising, and education to rapidly prototype visuals—such as custom art designs, brand-aligned product images, and teaching illustrations—while preserving intricate details, thereby lowering access to professional-grade content generation.

However, the ease of transferring visual identities raises prominent misuse concerns, including the generation of deceptive content like deepfakes for disinformation campaigns and unauthorized replication of copyrighted works that threaten intellectual property rights. Beyond ethical risks, reliance on large pre-trained diffusion models still causes significant energy consumption during inference—especially for high-resolution images—highlighting the need for efficient deployment research such as model compression.

To address these challenges, we advocate for ethical guidelines balancing creative freedom and anti-malicious safeguards. Specific measures include embedding imperceptible watermarks for traceability, implementing access controls for sensitive reference data like private portraits, and fostering industry collaboration to establish responsible use standards.

% \vspace{12pt}
% \color{red}
% IEEE conference templates contain guidance text for composing and formatting conference papers. Please ensure that all template text is removed from your conference paper prior to submission to the conference. Failure to remove the template text from your paper may result in your paper not being published.

\end{document}